\documentclass[10pt,twocolumn,letterpaper]{article}

\usepackage{cvpr}              

\usepackage{graphicx}

\usepackage[pagebackref,breaklinks,colorlinks,allcolors=cvprblue]{hyperref}
\usepackage{tikz}
\usepackage{adjustbox}
\usepackage{multirow}
\usepackage[table,xcdraw,dvipsnames]{xcolor}
\usepackage{tcolorbox}
\usepackage{graphicx}
\usepackage{booktabs}
\usepackage{float}
\usepackage{colortbl}
\usepackage{amsmath}
\usepackage{makecell}

\definecolor{datasetcolor}{HTML}{4472C4} 
\usepackage{tikz}
\definecolor{cvprblue}{rgb}{0.21,0.49,0.74}
\definecolor{firstb}{HTML}{FF6767}
\definecolor{thirdb}{HTML}{3564FF}
\definecolor{fovgreen}{HTML}{BED8CB}

\newcommand{\dscolor}[1]{\textcolor{datasetcolor}{#1}}
\newcommand{\dsname}{\textbf{\dscolor{\textsc{Denali}}}}
\newcommand{\underarrow}[2][black]{%
  \tikz[baseline=(text.base)]{
    \node[inner sep=0pt, outer sep=0pt] (text) {\textcolor{#1}{#2}};
    \draw[->, line width=0.5pt, color=#1] 
      ([yshift=0ex]text.south west) -- ([yshift=0ex]text.south east);
  }%
}
\newcommand{\fovhighlight}[2][fovgreen]{%
  \tikz[baseline=(text.base)]{
    \node[
      rectangle,
      rounded corners=2pt,
      fill=#1,
      inner ysep=0.2ex,
      inner xsep=0.4ex,
      outer sep=0pt
    ] (text) {\textcolor{black}{#2}};
  }%
}
\newcommand{\dsunderline}[1]{%
  \tikz[baseline=(text.base)]{
    \node[inner sep=0pt, outer sep=0pt, text=black] (text) {#1};
    \draw[line width=0.75pt, color=datasetcolor]
      ([yshift=-2.0pt]text.south west) -- ([yshift=-2.0pt]text.south east);
  }%
}
\newtcbox{\dsbutton}[1][]{%
  on line,
  arc=2mm,               
  colback=datasetcolor, 
  coltext=white!90!black,   
  boxrule=0pt,           
  boxsep=1pt,            
  left=3pt,
  right=3pt,
  top=1pt,
  bottom=1pt,
  nobeforeafter,
  tcbox raise base,
  #1
}
\newcommand{\bfheading}[1]{%
  \vspace{0.1em}%
  \noindent\textbf{#1}%
}

\usepackage[pagebackref,breaklinks,colorlinks,allcolors=cvprblue]{hyperref}

\def\paperID{35390}
\def\confName{CVPR}
\def\confYear{2026}

\title{%
\raisebox{-1.6ex}{\includegraphics[height=1.8em]{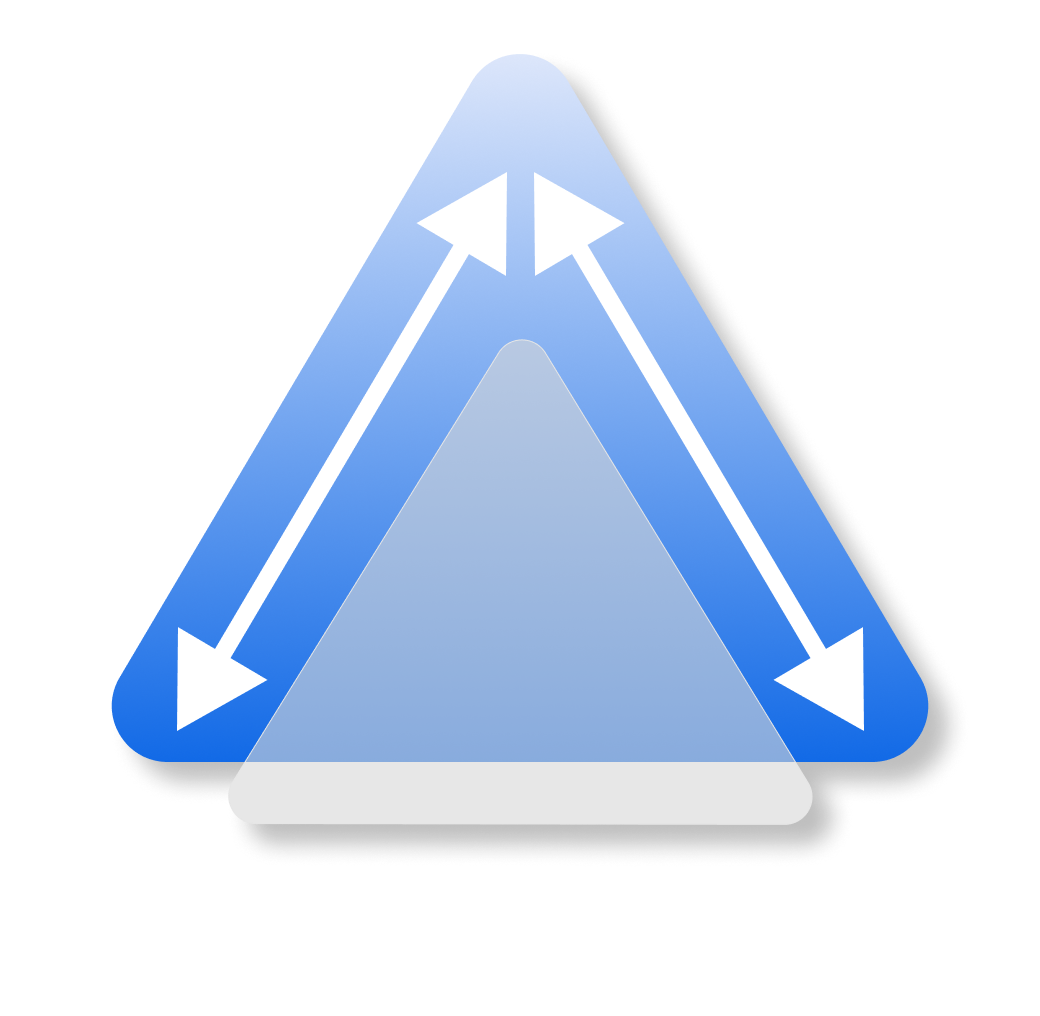}} \hspace{-0.2em}%
\dsname: A \dsunderline{D}ataset \dsunderline{E}nabling \dsunderline{N}on-Line-of-Sight \\ 
Sp\dsunderline{a}tial Reasoning with Low-Cost \dsunderline{Li}DARs%
}

\newcommand{\superscript}[1]{\ensuremath{^{\textrm{#1}}}}
\author{
    Nikhil Behari\superscript{1} \quad 
    Diego Rivero\superscript{1} \quad 
    Luke Apostolides\superscript{1} \\[0pt]
    Suman Ghosh\superscript{2,1} \quad 
    Paul Pu Liang\superscript{1} \quad 
    Ramesh Raskar\superscript{1}\\[3pt]
    \superscript{1}Massachusetts Institute of Technology \quad \superscript{2}Technische Universität Berlin \\[3pt]
    \href{https://nikhilbehari.github.io/denali}{\tt\small \color{magenta} nikhilbehari.github.io/denali}
    \vspace{-2pt}
}

\begin{document}

\twocolumn[{%
\renewcommand\twocolumn[1][]{#1}%
\maketitle
\begin{center}
    \centering
    \captionsetup{type=figure}
    \vspace{-2em}
    \includegraphics[width=\textwidth]{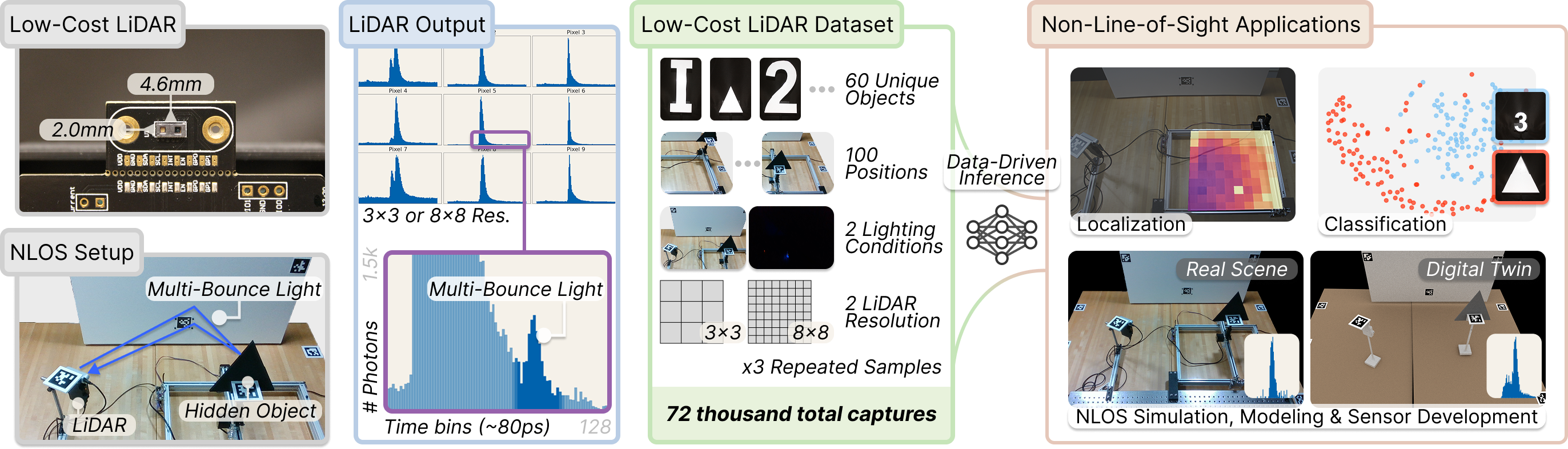} 
    \vspace{-1.5em}
    \captionof{figure}{\textbf{Enabling data-driven non-line-of-sight (NLOS) spatial reasoning with low-cost LiDARs.} \dsname{} is a large-scale dataset of 72{,}000 full time-resolved histograms captured with low-cost LiDARs in scenes designed to elicit multi-bounce returns from hidden objects. The dataset spans 60 object shapes, 100 positions, two lighting conditions, and two LiDAR resolutions. Using these diverse captures, we demonstrate the NLOS perception capabilities of consumer-grade LiDARs through data-driven methods, and identify key scene, modeling, and simulation-level limitations that can guide future work in data-driven NLOS inference.}
    \label{fig:teaser}
\end{center}%
}]

\maketitle

\begin{abstract}
Consumer LiDARs in mobile devices and robots typically output a single depth value per pixel. Yet internally, they record full time-resolved histograms containing direct and multi-bounce light returns; these multi-bounce returns encode rich non-line-of-sight (NLOS) cues that can enable perception of hidden objects in a scene. However, severe hardware limitations of consumer LiDARs make NLOS reconstruction with conventional methods difficult. In this work, we motivate a complementary direction: enabling NLOS perception with low-cost LiDARs through data-driven inference. We present DENALI, the first large-scale real-world dataset of space–time histograms from low-cost LiDARs capturing hidden objects. We capture time-resolved LiDAR histograms for 72,000 hidden-object scenes across diverse object shapes, positions, lighting conditions, and spatial resolutions. Using our dataset, we show that consumer LiDARs can enable accurate, data-driven NLOS perception. We further identify key scene and modeling factors that limit performance, as well as simulation-fidelity gaps that hinder current sim-to-real transfer, motivating future work toward scalable NLOS vision with consumer LiDARs.
\end{abstract}  
\vspace{-3.5em}
\section{Introduction}
\label{sec:intro}

Every time you take a photo\footnote{If you have an iPhone 12 Pro or newer}, your phone shoots out a grid of lasers in rapid succession. Silently, your phone is using light‑detection and ranging (LiDAR) to measure the depth of the scene; this LiDAR-based depth helps improve autofocus, portrait-mode separation, and augmented‑reality rendering. Beyond just mobile devices, these compact LiDAR sensors are increasingly used for depth sensing in robotics, AR/VR devices, and consumer electronics. 

Yet, LiDAR captures more than just scene depth: it also records time-resolved \textit{multi-bounce} light that reveals information about objects outside the sensor’s field of view. To measure depth, direct time-of-flight (dToF) LiDARs (hereafter \textit{LiDAR}) emit a short laser pulse into the scene; the sensor then measures the arrival times of returning photons using single-photon detectors with picosecond precision. These returns are accumulated into a \textit{histogram of photon-arrival times}, from which depth is estimated via the time bin of the dominant peak. Usually, each LiDAR histogram is reduced to this single depth value stored in a 3D point cloud. However, the full histogram contains far more signal beyond this peak, including \textit{late-arriving multi-bounce photons} that scattered off multiple (possibly hidden) surfaces before returning. These multi-bounce light returns can reveal subtle cues about hidden objects; this principle underlies non-line-of-sight (NLOS) imaging research, where multi-bounce LiDAR measurements have been used to recover the shape and reflectance of occluded objects. 


The advances in both LiDAR hardware and NLOS imaging prompt a natural question: \textit{if consumer LiDARs already record full photon-arrival histograms in everyday use, why is the informative multi-bounce signal discarded in favor of a single reported depth value?} A key reason is the mismatch between low-cost consumer LiDARs and lab-grade setups. Consumer dToF sensors are typically \textit{flash LiDARs} that illuminate the entire scene at once; they also have coarse spatial and temporal resolution, difficult-to-model crosstalk and noise~\cite{behari2025blurred,mu2024towards}, and are used in challenging real-world environments. In contrast, most existing NLOS methods rely on \textit{scanning LiDARs} with steered, collimated beams and high-timing-resolution single-photon detectors, usually evaluated in controlled laboratory settings. As a result, \textit{practical} NLOS imaging has not yet been demonstrated in consumer LiDAR hardware.

In this work, we motivate a new direction at this intersection: data-driven NLOS perception with low-cost LiDARs. Although these low-cost consumer LiDARs have severe hardware limitations, they already capture full time-resolved photon histograms, meaning multi-bounce light can be observed without laboratory-grade setups. While the spatial and temporal resolution constraints of these sensors make direct transfer of lab-based NLOS methods difficult, their scalability and deployment in real-world environments create an opportunity to learn NLOS perception directly from data. To advance this direction, we take a data-driven approach: quantifying the NLOS perception capabilities, and key limiting factors, of low-cost LiDARs using large-scale measurements in real scenes.

We introduce \dsname{}, the first large-scale real-world dataset capturing full histograms from low-cost LiDARs for non-line-of-sight perception. To assess the practical NLOS capabilities of these sensors, we construct diverse scenes with hidden geometries that elicit measurable three-bounce returns. Although the hardware constraints of these sensors preclude traditional NLOS \textit{reconstruction}, we show that their histograms enable a broader set of data-driven NLOS perception tasks, including object localization, shape classification, and size estimation. For each captured scene, we additionally generate a digital twin using a physically-based renderer. Using these real-sim pairs, we motivate applications of \dsname{} for advancing scalable NLOS perception with low-cost LiDARs, including improving simulation fidelity, sim-to-real transfer, and task-aware sensor design.

\noindent \textbf{Our contributions in this work are as follows: }

\begin{itemize}
    \item  We introduce \dsname{}, the first large-scale real-world dataset using low-cost LiDARs to capture full-histogram signal for data-driven non-line-of-sight perception. 
    \item Using \dsname{}, we demonstrate multiple forms of data-driven NLOS perception achievable with low-cost LiDARs; we also identify key \textit{scene and modeling-dependent factors} that limit performance, motivating future work in robust data-driven NLOS inference.
    \item As part of \dsname{}, we generate digital twins for each scene capture in a physically-based simulator; we demonstrate how this paired data enables quantitative analysis of NLOS simulation fidelity, sim-to-real transfer, and LiDAR design for NLOS perception.
\end{itemize}

\bfheading{Scope of this Work.}
We capture data using low-cost LiDARs in real-world but controlled conditions. Scenes are designed to elicit multi-bounce returns: objects are retroreflective, constrained within a known bounding region, and viewed from fixed sensor, table, and relay-wall. We ensure capture consistency to better characterize these sensors in best-case conditions; however, this does not directly reflect the variability of dynamic real-world environments. Extending this work to unconstrained scenes remains an open direction. Our data is captured with a single LiDAR model (ams TMF8828 \cite{ams_osram_tmf882x}), representative of, but not identical to, a broader class of compact dToF sensors (\cite{st_VL53L8CX}).
\section{Related Work}
\label{sec:relwork}


\bfheading{Perception with Time-of-Flight Sensing.} LiDAR is widely used in autonomous driving, robotics, and mobile imaging, often alongside RGB cameras to provide robust metric depth in challenging environments \cite{roriz2021automotive,yue2024lidar,huang2022multi}. Conventional spinning and solid-state dToF LiDARs estimate depth using only the first or strongest photon return from each emitted pulse, producing a single depth value per pixel that is typically stored in a 3D point cloud \cite{behroozpour2017lidar}. Point-cloud LiDAR data has driven major advances in 3D perception \cite{qian20223d,lang2019pointpillars,weng2020ab3dmot,behley2019semantickitti}, supported by large-scale datasets such as KITTI \cite{geiger2012we}, nuScenes \cite{caesar2020nuscenes}, Waymo Open \cite{sun2020scalability}, ARKitScenes \cite{baruch2021arkitscenes}, and SemanticKITTI \cite{behley2019semantickitti}. Each of these datasets treat LiDAR measurements as point depth estimates. However, dToF LiDARs inherently measure richer scene information: by measuring the full temporal distribution of returning photons (i.e. \textit{beyond} the strongest returns), LiDARs also capture rich multi-bounce light transport in a scene. 

\textit{Transient imaging} refers to the use of these time-resolved light measurements, using full LiDAR histograms (\textit{transients}) for richer scene understanding \cite{piron2020review,pu2025computational}. Prior work has shown that this histogram signal can be used to see through scattering media \cite{satat2016all,satat2018towards}, estimate material properties \cite{su2016material,callenberg2021low}, and disambiguate complex (e.g. glossy, semi-transparent) scene geometries \cite{naik2015light,lin2024handheld}. One well-known use of full-histogram returns is for non-line-of-sight (NLOS) imaging, where late-arriving multi-bounce photons are used to estimate the geometry of hidden scene regions \cite{kirmani2009looking,velten2012recovering,isogawa2020optical,faccio2020non}. Several recent methods have aimed to leverage neural network-based approaches for NLOS reconstruction \cite{shen2021non,fujimura2023nlos}; in parallel, recent work has explored learning data-driven representations for NLOS imaging \cite{tancik2018data,chen2020learned,li2023nlost,li2024toward}. Critically, however, these methods primarily focus on \textit{reconstructing} hidden scenes with \textit{laboratory-grade LiDAR setups}. In this work, we uniquely aim to capture and quantify the broader NLOS perception capabilities of deployable, \textit{low-cost consumer LiDARs}. 

\bfheading{Imaging with Low-Cost LiDARs.} Advances in consumer hardware have enabled time-of-flight (ToF) sensing in low-cost, consumer devices (e.g. ams TMF8828 \cite{ams_osram_tmf882x} and ST VL53L8CX \cite{st_VL53L8CX}). Recent works have demonstrated that these low-cost consumer LiDARs can be used in isolated settings for depth sensing, material classification, and NLOS vision \cite{callenberg2021low}; similarly, recent works have shown that these low-cost LiDARs can be used on mobile robots to improve hidden obstacle avoidance with NLOS vision \cite{young_enhancing_2025}. In parallel, several methods have begun to exploit the full temporal histogram of these sensors for 3D reconstruction \cite{behari2025blurred,mu2024towards}, robotic planning \cite{sifferman2025efficient}, and scene understanding \cite{sifferman2025recovering}. Despite these advances, the few existing works that explore NLOS vision with low-cost LiDARs operate in small-scale, narrowly controlled conditions. In contrast, we introduce a large-scale real-world dataset designed to characterize and benchmark the NLOS capabilities of these sensors across diverse objects, poses, and capture settings.
\section{Dataset}
\label{sec:dataset}

\begin{figure}[!tb]
\centering
\includegraphics[width=\linewidth]{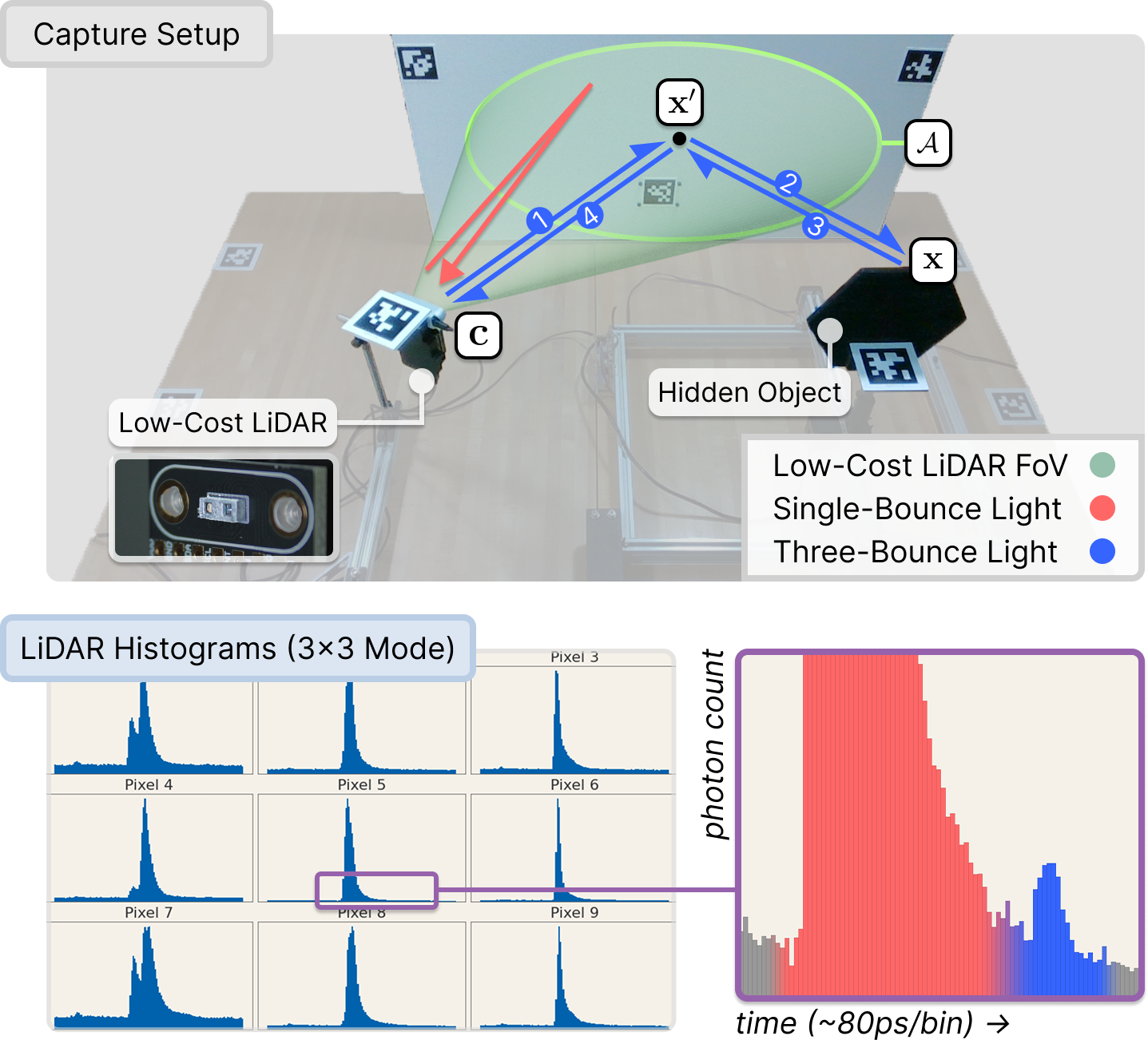}
\caption{\textbf{Non-line-of-sight signal in mobile LiDARs.} Conventional direct time-of-flight (dToF) LiDARs report depth measurements corresponding to primary single-bounce reflections (shown in \textcolor{firstb}{red}). However, dToF LiDARs can also capture later-arriving multi-bounce photons (shown in \textcolor{thirdb}{blue}); this \textit{three-bounce light signal} can encode information about hidden scene objects which may be used for non-line-of-sight inference.}
\label{fig:nlos_explain}
\vspace{-1.5em}
\end{figure}

\begin{figure*}[!ht]
  \centering
  \includegraphics[width=1.0\textwidth]{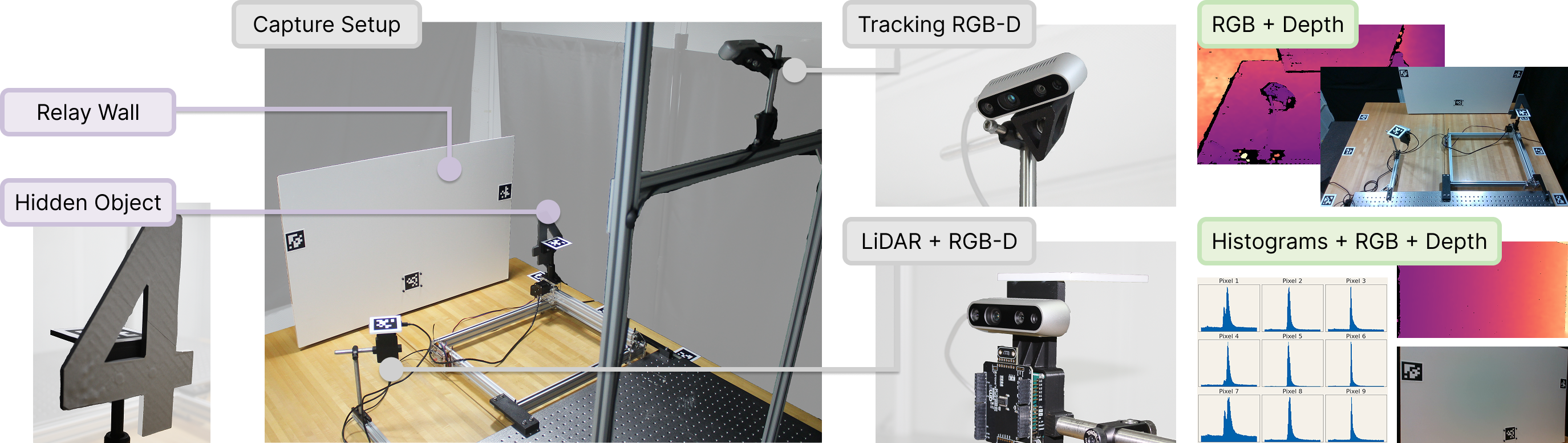}
    \vspace{-6mm}
   \caption{\textbf{Capture setup.} Our capture system is designed to record a large-scale dataset of non-line-of-sight, three-bounce light signals from a mobile flash LiDAR. The setup includes a low-cost single-photon LiDAR co-located with an Intel RealSense RGB-D camera, both directed toward a flat relay wall. A hidden object is mounted on a motorized gantry positioned outside their direct line of sight to ensure only indirect three-bounce returns are measured. An additional overhead RealSense tracking camera observes the entire scene, providing accurate localization of the LiDAR, relay wall, and hidden object during data collection.}
  \label{fig:capture_setup}
  \vspace{-1em}
\end{figure*}

We next describe our dataset acquisition process. We first explain how low-cost consumer LiDARs can ``see" hidden objects using multi-bounce light returns, and describe the specific LiDAR used in this work. We then describe our capture setup, designed to emphasize three-bounce light, \textit{and} describe our digital twin generation for each captured scene. Finally, we summarize the complete dataset.

\subsection{Three-Bounce Light in Consumer LiDARs}
\label{subsec:lidar_used}
DToF LiDARs, including those used in mobile devices, estimate depth by emitting short laser pulses and recording the return times of detected photons. Each LiDAR pixel performs this measurement repeatedly, accumulating photon counts into a \textit{temporal histogram} that records photon returns over time.  The time $t$ of the first and strongest peak corresponds to light that travels directly to a visible surface and back, yielding a depth estimate $\frac{ct}{2}$ for $c$ the speed of light (\underarrow[firstb]{red path} in \cref{fig:nlos_explain}). However, late-arriving returns, while weaker, can also capture multiply scattered photons. In particular, we focus on \textit{three-bounce light returns}, consisting of photons reflected from a visible surface to a hidden object, back to the visible surface, and then to the detector (\underarrow[thirdb]{blue path} in \cref{fig:nlos_explain}); these signals encode information about occluded geometries, enabling NLOS perception.

Let $\mathbf{c}$ denote the LiDAR position, $\mathbf{x}'$ a point on the visible ``relay wall", and $\mathbf{x}$ a point on the hidden object. In a typical \textit{confocal} configuration, a co-located collimated laser and single-photon detector are pointed at the same spot $\mathbf{x}'$ on the visible relay wall. Three-bounce NLOS light then follows the path $\mathbf{c} \rightarrow \mathbf{x}' \rightarrow \mathbf{x} \rightarrow \mathbf{x}' \rightarrow \mathbf{c}$, traveling from the laser to the wall, then to the hidden object, back to the wall, and finally returning to the detector. Since the camera-wall distance is known from direct LiDAR depth measurements, it is common to express the NLOS measurement observed at relay-wall point $\mathbf{x}'$ with respect to the hidden scene geometry. The histogram response at each point on the relay wall $\tau(\mathbf{x}', t)$ can then be written as: 

\vspace{-3mm}
\begin{equation}
\tau(\mathbf{x}', t) = \iiint_{\Omega} \rho(\mathbf{x}) 
\frac{\delta\!\left(2\|\mathbf{x}' - \mathbf{x}\| - c t\right)}
{\|\mathbf{x}' - \mathbf{x}\|^{4}} \, d\mathbf{x},
\label{eq:histogram_response}
\end{equation}

\noindent where $\Omega$ denotes the hidden volume, $\rho(\mathbf{x})$ is the albedo of the hidden surface, $c$ is the speed of light, and $\delta(\cdot)$ enforces that only points whose round-trip distance $2\|\mathbf{x}' - \mathbf{x}\|$ equals the observed time-of-flight $c t$ contribute at time $t$. We visualize an example of first and third-bounce histogram returns in \cref{fig:nlos_explain}.

\vspace{0.5mm}
\noindent \bfheading{Low-Cost LiDAR Used for Data Capture.} We capture LiDAR measurements using the ams TMF8828, a 940 nm consumer-grade flash dToF sensor with characteristics comparable to those in modern mobile devices \cite{ams_osram_tmf882x}. This low-cost module ($\sim$\$10) reports full photon-arrival histograms over 128 discrete time bins, operating in either $3 \times 3$ or $8 \times 8$ pixel output modes. We capture each of our scenes in both modes: 8×8 outputs provide finer spatial sampling but records significantly fewer photons per pixel. Photon timing is measured using an integrated SPAD array and on-chip time-to-digital converter (TDC). The module supports short-range ($\approx$1.5m) and long-range ($\approx$5m) depth modes; we use the short-range mode for all captures.

Unlike the idealized formulation in \cref{eq:histogram_response}, this low-cost LiDAR emits a flood-illumination rather than a collimated laser beam; each detector pixel also integrates light over a wide instantaneous field of view (iFoV) rather than a narrow scene point (example \fovhighlight{wide FoV} shown in \cref{fig:nlos_explain}). We assume that our hidden object is retroreflective, which preferentially returns light along the incoming illumination direction; under this assumption, the temporal histogram measured by pixel $p$ can be expressed as a weighted sum of contributions from wall points within its iFoV:
\begin{equation}
    \tau_p(t) = \int_{\mathcal{A}_p} w_p(\mathbf{x}') \, \tau(\mathbf{x}', t) \, d\mathbf{x}',
\end{equation}
where $\mathcal{A}_p$ denotes the region of the relay wall imaged by pixel $p$, $w_p(\mathbf{x}')$ encodes the pixel’s spatial sensitivity over that region, and $\tau(\mathbf{x}', t)$ is the confocal transient response at wall point $\mathbf{x}'$.

\subsection{Capture Setup}
We next describe our dataset capture setup, designed to elicit measurable three-bounce returns from hidden objects.

\begin{figure*}[!ht]
  \centering
    \vspace{0mm}
    \includegraphics[width=1.0\textwidth]{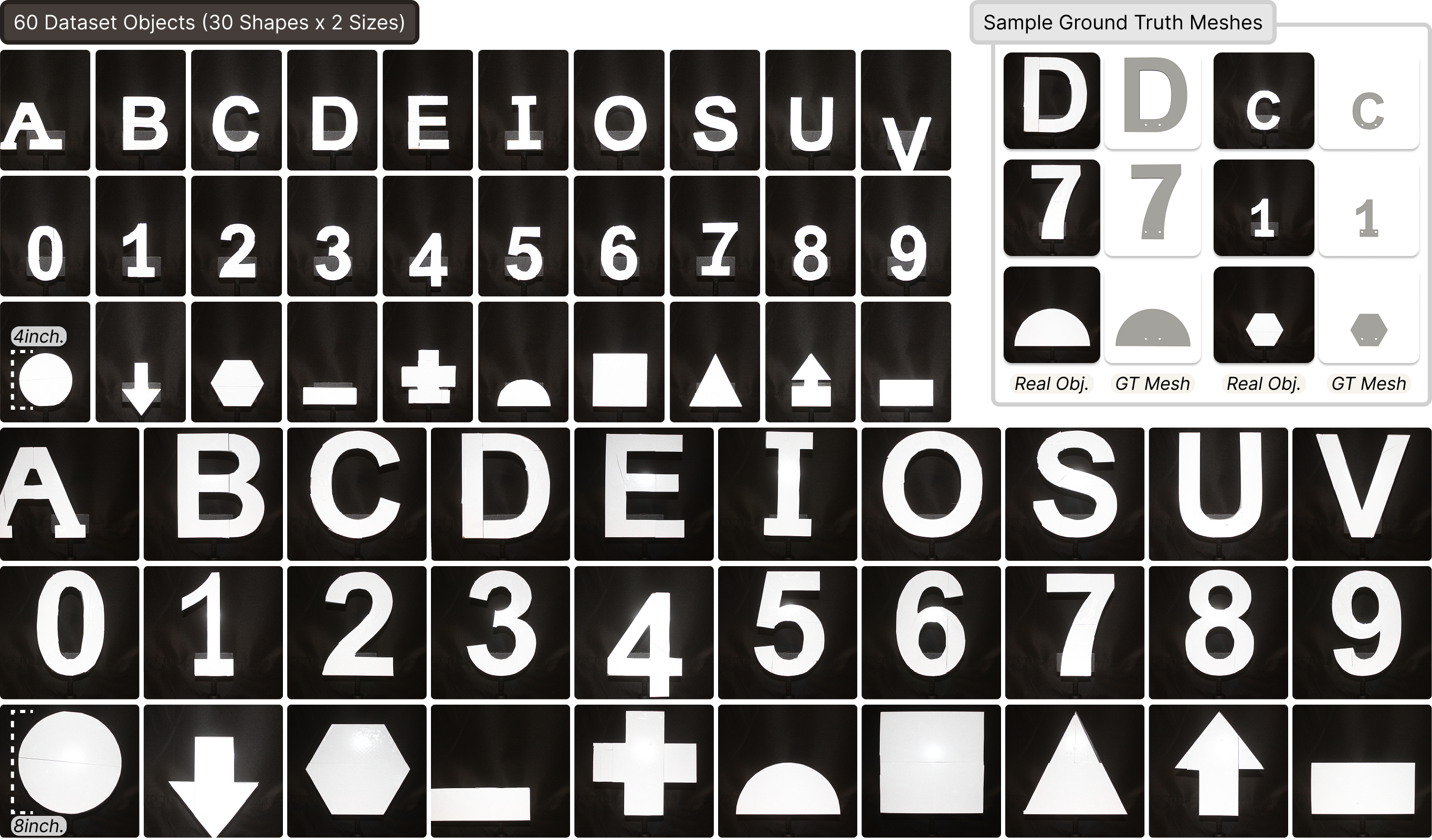}
    \vspace{-6mm}
    \caption{\textbf{Captured dataset objects.} We 3D print 30 objects (10 letters, 10 numbers, 10 shapes) at two scales (4in. and 8in.). Retroreflective tape is applied to improve three-bounce photon returns in the measured LiDAR histograms. Each object is printed from a known CAD model; as such, accurate ground-truth meshes are available for downstream analysis and simulation.}
  \label{fig:dataset_objects}
  \vspace{-4mm}
\end{figure*}

\bfheading{Scene Configuration.} Our objective is to capture a large-scale dataset containing measurable three-bounce returns in low-cost LiDAR histograms. To do so, we orient the LiDAR such that its FoV primarily observes a flat vertical relay wall. This wall acts as the intermediary surface that directs illumination toward the hidden object and redirects its returning photons back to the sensor, creating the canonical three-bounce NLOS path.

We mount objects on a motorized gantry located outside the direct line of sight of the LiDAR; this ensures that LiDAR measurements of the objects arise solely from three-bounce returns. The gantry samples 100 object positions in the ground-plane $(x,y)$ directions, with each location fully outside of the sensor FoV. Our full capture setup is shown in \cref{fig:capture_setup}. We 3D-print 30 distinct objects in two bounding-box sizes (4in. and 8in.), shown in \cref{fig:dataset_objects}, each with known geometry for later use in simulation and validation. Every object includes two mounting holes for attachment to a custom 3D-printed fixture that defines a known transformation between the object, its mount, and an attached tracking tag (\cref{subsec:poses}). To enhance the three-bounce signal from hidden objects in the measured histograms, we apply retroreflective tape to the object surfaces (see \cref{fig:dataset_objects}).

\bfheading{Scene Tracking.} Our capture setup includes three synchronized sensors for accurate global tracking. We capture LiDAR histogram data with the TMF8828 flash LiDAR as described in \cref{subsec:lidar_used}. Adjacent to the LiDAR, we place a \textit{co-located} Intel RealSense D435i \cite{intel_realsense} that captures 848×480 RGB and aligned depth (RGB-D) images derived from stereo infrared imaging. The TMF8828 and co-located RealSense are secured on a custom 3D-printed rigid mount that maintains fixed relative alignment between their FoVs; the mount has a known geometry which supports downstream geometric calibration and simulation. We additionally capture the full scene from a \textit{tracking} RealSense D435i camera positioned above the scene. The tracking camera captures 1280×720 RGB-D data covering the LiDAR sensor, relay wall, and hidden object. We show this setup and sample outputs from these sensors in \cref{fig:capture_setup}.

\subsection{Digital Twin Capture}
\label{subsec:poses}
Our dataset includes 6-DoF ground truth poses of the LiDAR, hidden objects, relay wall, and tabletop surface that we use to generate digital twins for every captured scene. To localize scene features, we attach AprilTag \cite{tinkerTwins_apriltag_2023} markers (size 6cm, tag36h11 family) in known positions on the table (IDs 10-15), relay wall (IDs 5-7), LiDAR (ID 0) and the hidden object (ID 1). We estimate poses for each tag across $\sim$12,400 light-on captures; we then remove outlier detections with $|z|>2$, and compute the mean pose for each marker. We include a full analysis of quantified pose prediction accuracy and calibration details in the supplement.  
 
Using these calibrated poses, we reconstruct a digital twin for every captured scene in Mitsuba 3 \cite{jakob2022mitsuba3}. We generate each digital twin by combining estimated tag poses with the known rigid transforms between tags and scene elements (LiDAR, hidden object, relay wall, table plane), yielding a complete 3D scene geometry, including the ground-truth object mesh at its calibrated pose. We illustrate digital-twin pairs for sample captures in \cref{fig:teaser} and \cref{fig:simulated_scene}. We use the co-located and tracking RGB streams for tag localization; RealSense depth is not used in our digital-twin generation, but is provided for additional validation or refinement. Further details on the scene parameterization and our rendering approach are included in the supplement.

\begin{figure*}[ht]
  \centering
  \begin{minipage}[t]{0.21\textwidth}
    \vspace{0pt}
    \centering
    \footnotesize
    \resizebox{\textwidth}{!}{%
      \begin{tabular}{p{2.0cm} c}
        \toprule
        \textbf{Dimension} & \textbf{Count} \\
        \midrule
        Unique objects          & 60 \\
        Object positions        & 100 \\
        LiDAR pixels            & 3x3 / 8x8 \\
        Lighting \quad \quad conditions     & Lights on / off \\
        Repeated \quad \quad samples        & 3x samples \\
        Histogram bins          & 128 \\
        \midrule
        \textbf{Total captures}  & \textbf{72{,}000} \\
        \textbf{Total pixels}  & \textbf{2{,}628{,}000} \\
        \textbf{Total ToF bins}  & \textbf{336{,}384{,}000} \\
        \bottomrule
      \end{tabular}%
    }
  \end{minipage}
  \hfill
  \begin{minipage}[t]{0.77\textwidth}
    \vspace{0pt}
    \centering
    \includegraphics[width=\textwidth]{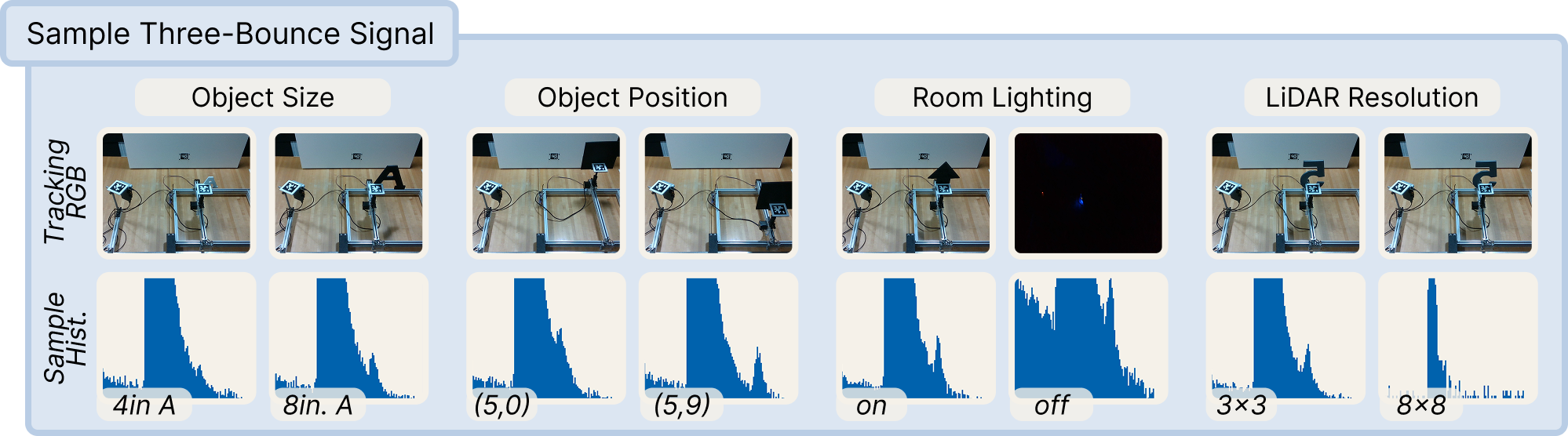}
  \end{minipage}
  \caption{\textbf{Summary of dataset captures (left) and sample three-bounce signals from low-cost LiDAR (right).} Example histograms from the central pixel of \(3 \times 3\) and \(8 \times 8\) configurations illustrate variations in three-bounce signal: intensity differences by object size, temporal shifts by position, and changes in photon counts and noise across lighting conditions and LiDAR resolution.}
  \label{fig:dataset_summary_figure}
  \vspace{-1em}
\end{figure*}

\subsection{Summary of Captures}
Our full collected dataset contains:
60 objects $\times$ 100 locations $\times$ 2 LiDAR spatial resolutions $\times$ 2 lighting conditions $\times$ 3 repeated samples. This amounts to 72,000 total LiDAR captures, consisting of 2,628,000 full-histogram LiDAR pixels captured or 336,384,000 distinct time-of-flight bin measurements. Each capture is also paired with a corresponding digital twin rendered in Mitsuba 3. We summarize the dataset and visualize representative three-bounce measurements across capture conditions in \cref{fig:dataset_summary_figure}. Below, in \cref{tab:hist_stats}, we report a statistical analysis of the measured histograms after background subtraction, which isolates the captured three-bounce signal.

\vspace{-5pt}
\begin{table}[h]
\centering
\caption{\textbf{Statistical analysis of three-bounce signal.} For each capture, we subtract the corresponding no-object background to isolate the three-bounce return. We report mean $\pm$ SEM for pixels [1,1] ($3\times3$) and [2,1] ($8\times8$).}
\vspace{-4pt}
\resizebox{\columnwidth}{!}{
\begin{tabular}{llccccc}
\toprule
\textbf{Res.} & \textbf{Light} & \textbf{Size} &
\textbf{Total Intensity} &
\textbf{Centroid (bins)} &
\textbf{Spread (bins)} &
\textbf{Skew} \\
\midrule
\multirow{4}{*}{3$\times$3}
 & \multirow{2}{*}{On}  & 4in. & 560.36 ± 6.08 & 91.55 ± 0.08 & 12.25 ± 0.05 & 0.40 ± 0.01 \\
 &                       & 8in. & 1468.33 ± 14.31 & 96.40 ± 0.07 & 8.67 ± 0.05 & -0.09 ± 0.02 \\
\cmidrule(l){2-7}
 & \multirow{2}{*}{Off} & 4in. & 1612.88 ± 11.08 & 94.64 ± 0.05 & 14.73 ± 0.03 & 0.23 ± 0.01 \\
 &                       & 8in. & 2448.56 ± 16.27 & 96.57 ± 0.05 & 12.04 ± 0.04 & 0.12 ± 0.01 \\
\midrule
\multirow{4}{*}{8$\times$8}
 & \multirow{2}{*}{On}  & 4in. & 11.72 ± 0.08 & 94.74 ± 0.07 & 15.93 ± 0.03 & 0.19 ± 0.01 \\
 &                       & 8in. & 11.99 ± 0.08 & 94.79 ± 0.06 & 15.85 ± 0.03 & 0.18 ± 0.01 \\
\cmidrule(l){2-7}
 & \multirow{2}{*}{Off} & 4in. & 18.69 ± 0.06 & 97.45 ± 0.04 & 16.47 ± 0.02 & 0.07 ± 0.00 \\
 &                       & 8in. & 19.04 ± 0.06 & 97.46 ± 0.04 & 16.40 ± 0.02 & 0.07 ± 0.00 \\
\bottomrule
\end{tabular}}
\vspace{-0.5em}
\label{tab:hist_stats}
\end{table}

\vspace{-0.2em} 
\section{Analysis}
\label{sec:analysis}
\vspace{-1mm}
We have three primary analysis goals: (1) to characterize and quantify the NLOS capabilities of low-cost LiDARs using data-driven methods, (2) to determine which model architectures are most effective for NLOS perception, and (3) to highlight potential weaknesses in how current modeling techniques leverage histogram data for NLOS inference. To this end, we evaluate three tasks using our dataset:

\vspace{0.6em}
\noindent \dsbutton{\textsc{Denali} tasks}
\begin{enumerate}
    \item \underline{NLOS Object Localization}: Predicting the continuous planar \((x, y)\) position of an occluded object.
    \item \underline{NLOS Object Classification}: Identifying the object's shape among 30 discrete categories (across object sizes).
    \item \underline{NLOS Size Classification}: Determining whether the object corresponds to a 4-inch or 8-inch physical scale.
\end{enumerate}
\vspace{0.3em}

\noindent For each task, the input is a LiDAR photon-count tensor of shape \((n, n, 128)\), where \(n\) denotes the spatial resolution and 128 denotes the temporal histogram bins. For the following analysis, \textit{we focus specifically on \(3 \times 3\) LiDAR captures}; we include full results on inference using \(8 \times 8\) LiDAR resolution in the supplement.

\bfheading{Models and Evaluation.}  
We evaluate four models that differ in how they use spatial and temporal structure in the LiDAR histograms. Each architecture introduces a distinct inductive bias (i.e. from structure-agnostic to fully spatiotemporal), allowing us to benchmark how modeling choices affect data-driven NLOS inference:

\vspace{0.6em}
\noindent \dsbutton{Benchmarked models}
\begin{enumerate}
    \item \underline{Baseline MLP}: Each \(3 \times 3 \times 128\) capture is flattened and passed through a fully connected network; baseline with no spatial or temporal inductive bias.
    
    \item \underline{1D CNN (Time-Only)}: Pixels are stacked as channels; 1D convolutions operate along time bins, using temporal structure without spatial inductive bias.
    
    \item \underline{3D CNN (Spatiotemporal)}: Convolutions operate jointly over time bins and the \(3 \times 3\) spatial grid; captures local spatiotemporal features before global pooling.
    
    \item \underline{Transformer (Time-Token Encoder)}: Each time bin is treated as a token with positional encoding; captures long-range temporal dependencies via self-attention (with no spatial inductive bias).
\end{enumerate}
\vspace{0.6mm}

\noindent For each task, we train a supervised model with a task-appropriate loss (mean squared error for localization, categorical cross-entropy for classification, binary cross-entropy for size prediction). All \(3 \times 3\) LiDAR samples (across object size, position, lighting condition on/off, and repeated captures) are randomly split 70/30 into train and test sets. Metrics are reported on the held-out test set. Full model and training details are included in the supplement.

\begin{figure}[b]
\centering
\vspace{-3mm}
\includegraphics[width=\linewidth]{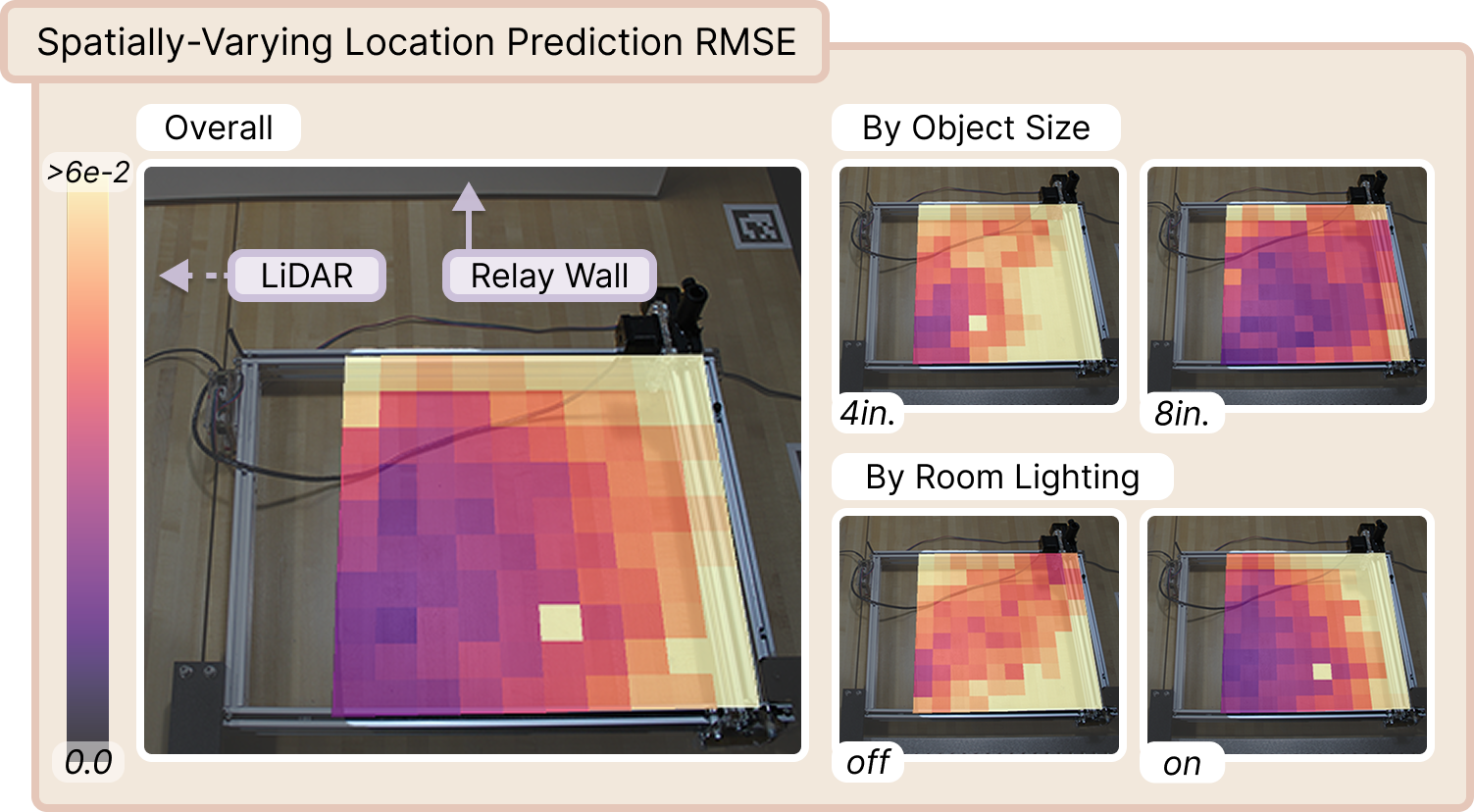}
\vspace{-6mm}
\caption{\textbf{Spatial mapping of NLOS localization accuracy.} We plot RMSE (m) over true gantry positions for a single trained model (1D CNN) broken down by size/lighting. Accuracy generally improves for larger (8in.) objects nearer to relay wall. However, different lighting induces distinct spatial error patterns, suggesting poor separation of object, geometry, and lighting.}
\label{fig:object_loc_heatmap}
\end{figure}

\begin{table*}[!ht]
\centering
\caption{\textbf{Benchmarking NLOS perception tasks using low-cost LiDARs with \dsname{}.}
We report performance for (a) location regression (RMSE / MAE), (b) object classification (Top-1 / Top-5 / Macro-F1), and (c) size prediction (Precision / Recall / Accuracy). Results are reported overall and by scene factors; CNN-based models perform best but remain dependent on object size, position, and shape.}
\vspace{-3mm}
\label{table:nlos_benchmarks}
\tiny
\resizebox{0.95\textwidth}{!}{
\begin{tabular}{llcccc}
\toprule
\textbf{Metric Group} & \textbf{Subgroup} & \textbf{MLP} & \textbf{1D CNN} & \textbf{3D CNN} & \textbf{Transformer} \\
\midrule
\rowcolor{lightgray!50}\multicolumn{6}{l}{\textbf{(a) Location Regression (RMSE $\downarrow$ / MAE $\downarrow$ )}} \\
\midrule
Overall & - & 0.1045 / 0.0907 & 0.0456 / 0.0324 & 0.0475 / 0.0337 & 0.0579 / 0.0428 \\
\midrule
\multirow{2}{*}{By Size} & \hspace{0.5em}4 inches & 0.1044 / 0.0907 & 0.0548 / 0.0398 & 0.0537 / 0.0408 & 0.0678 / 0.0522 \\
                         & \hspace{0.5em}8 inches & 0.1046 / 0.0907 & 0.0341 / 0.0250 & 0.0404 / 0.0266 & 0.0462 / 0.0335 \\
\midrule
\multirow{3}{*}{By Object Class} & \hspace{0.5em}Numbers & 0.1044 / 0.0907 & 0.0497 / 0.0334 & 0.0477 / 0.0341 & 0.0593 / 0.0439 \\
                                 & \hspace{0.5em}Letters & 0.1043 / 0.0903 & 0.0450 / 0.0335 & 0.0508 / 0.0346 & 0.0595 / 0.0444 \\
                                 & \hspace{0.5em}Shapes  & 0.1049 / 0.0911 & 0.0414 / 0.0301 & 0.0431 / 0.0322 & 0.0551 / 0.0403 \\
\midrule
\rowcolor{lightgray!50}\multicolumn{6}{l}{\textbf{(b) Object Classification (Top-1 $\uparrow$ / Top-5 $\uparrow$ / Macro-F1 $\uparrow$)}} \\
\midrule
Overall & - & 0.0665 / 0.3062 / 0.0389 & 0.3876 / 0.7954 / 0.3832
& 0.3523 / 0.5737 / 0.4377 & 0.1167 / 0.4019 / 0.1003 \\
\midrule
\multirow{2}{*}{By Location} & \hspace{0.5em}Near quadrant       & 0.0538 / 0.2586 / 0.0280 & 0.4075 / 0.7787 / 0.3695 & 0.3476 / 0.5545 / 0.3883 & 0.1101 / 0.3698 / 0.0811 \\
                             & \hspace{0.5em}Far quadrant & 0.0741 / 0.3197 / 0.0362 & 0.3826 / 0.7712 / 0.3463 & 0.3624 / 0.5631 / 0.4059 & 0.1187 / 0.4156 / 0.0904 \\
\midrule
\multirow{2}{*}{By Size} & \hspace{0.5em}4 inches & 0.0796 / 0.3239 / 0.0460 & 0.3191 / 0.7306 / 0.2853 & 0.3470 / 0.5559 / 0.4323 & 0.1199 / 0.4171 / 0.1061 \\
                         & \hspace{0.5em}8 inches & 0.0537 / 0.2646 / 0.0300 & 0.4573 / 0.8213 / 0.4397 & 0.3577 / 0.5643 / 0.4386 & 0.1136 / 0.3833 / 0.0852 \\
\midrule
\rowcolor{lightgray!50}\multicolumn{6}{l}{\textbf{(c) Size Prediction (Precision $\uparrow$ / Recall $\uparrow$ / Accuracy $\uparrow$)}} \\
\midrule
Overall & - & 0.5968 / 0.5363 / 0.5363 & 0.9488 / 0.9468 / 0.9468 & 0.9304 / 0.9298 / 0.9298 & 0.8727 / 0.8722 / 0.8722 \\
\midrule
\multirow{2}{*}{By Location} & \hspace{0.5em}Near quadrant & 0.5675 / 0.5254 / 0.5217 & 0.9627 / 0.9599 / 0.9603 & 0.9569 / 0.9555 / 0.9559 & 0.9157 / 0.9075 / 0.9078 \\
                             & \hspace{0.5em}Far quadrant & 0.5973 / 0.5482 / 0.5524 & 0.9447 / 0.9444 / 0.9433 & 0.9276 / 0.9264 / 0.9255 & 0.8756 / 0.8619 / 0.8591 \\
\midrule
\multirow{3}{*}{By Object Class} & \hspace{0.5em}Numbers & 0.5366 / 0.5202 / 0.5125 & 0.9570 / 0.9544 / 0.9543 & 0.9482 / 0.9419 / 0.9415 & 0.9050 / 0.8938 / 0.8920 \\
                                 & \hspace{0.5em}Letters & 0.6600 / 0.5357 / 0.5427 & 0.9627 / 0.9591 / 0.9593 & 0.9381 / 0.9286 / 0.9288 & 0.8713 / 0.8595 / 0.8599 \\
                                 & \hspace{0.5em}Shapes  & 0.5847 / 0.5534 / 0.5539 & 0.9392 / 0.9258 / 0.9266 & 0.9310 / 0.9182 / 0.9190 & 0.8928 / 0.8638 / 0.8649 \\
\toprule
\end{tabular}
}
\vspace{-4mm}
\end{table*}

\begin{figure}[t]
\centering
\includegraphics[width=\linewidth]{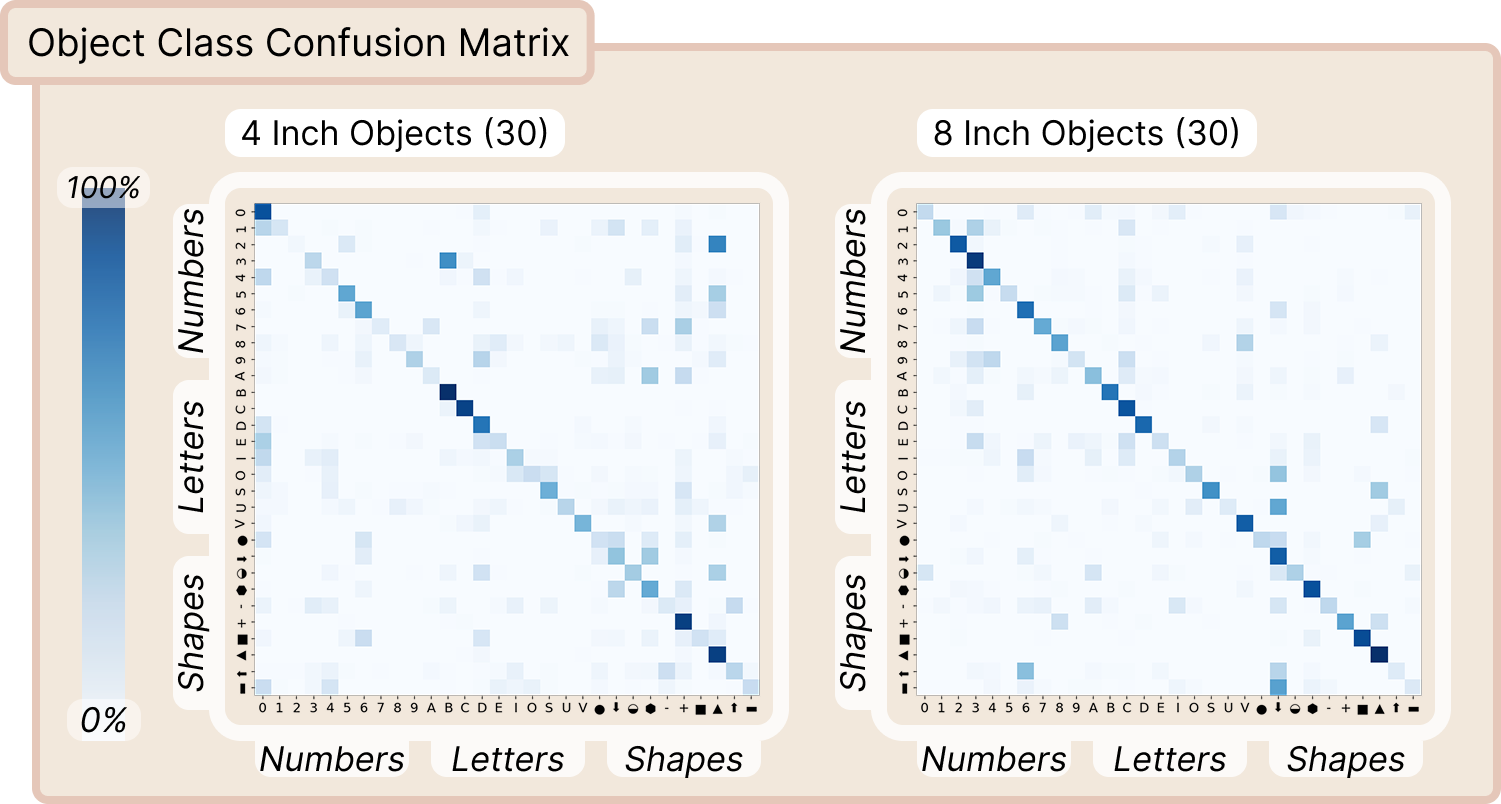}
\vspace{-6mm}
\caption{\textbf{NLOS classification accuracy across object types.} We train a 1D CNN independently for each object size and evaluate classification performance across shapes; larger 8-inch objects yield consistently higher accuracy.}
\label{fig:object_class_confusion}
\vspace{-1.5em}
\end{figure}

\bfheading{Results and Discussion.} We benchmark all model architectures across tasks in \cref{table:nlos_benchmarks}. Overall, we find that histogram signal from low-cost LiDAR supports accurate NLOS localization, shape classification, and size prediction. Our best-performing models achieve an RMSE of 0.046m for localization (shown in \cref{fig:object_loc_heatmap}), a macro-F1 of 0.38 for classification, and an accuracy of 0.95 for size prediction. These results indicate that low-cost consumer LiDAR histograms contain sufficient information to enable a range of NLOS perception tasks, suggesting a feasible path toward deployable NLOS applications beyond traditional reconstruction.

Convolutional architectures (e.g., 1D and 3D CNNs) consistently achieve the best performance across tasks (\cref{table:nlos_benchmarks}), indicating that an inductive bias toward local temporal structure is well suited for our LiDAR histogram data. Notably, the 3D CNN (which can exploit spatial cues across pixels) did not outperform the 1D CNN; this suggests that current models struggle to utilize the available, but low-resolution, spatial information in low-cost LiDARs.

We observe two potential weaknesses in data-driven perception with low-cost LiDAR, one scene-dependent and one modeling-dependent. At the scene level, NLOS perception with low-cost LiDAR depends strongly on both object size and location. Larger objects are significantly easier to detect in histograms: \cref{fig:object_loc_heatmap} shows that 8-inch objects can be localized over a wider spatial region than 4-inch objects, and \cref{fig:object_class_confusion} shows consistently higher classification accuracy for 8-inch objects. Object location is equally important: objects closer to the relay wall enable better NLOS perception (\cref{table:nlos_benchmarks}), but objects placed \textit{too} close to the wall cannot be localized due to overlapping first-bounce and three-bounce returns (\cref{fig:object_loc_heatmap}). While scene factors (size, location) influence both conventional and NLOS perception, our results show that their effects on NLOS perception are distinct, motivating a clearer mapping of these scene-based limitations for NLOS from low-cost LiDARs. 

At the modeling level, \cref{fig:object_loc_heatmap} reveals unexpectedly different spatial error patterns across lighting conditions for a single trained model (we would expect that global illumination affects error \textit{uniformly}). This indicates that current models may not cleanly disentangle object properties, scene geometry, and ambient illumination in histogram data. Developing models that better factorize object, geometry, and lighting-related effects remains an important direction for robust NLOS perception with low-cost LiDARs.

\bfheading{Generalization.} To test whether these findings extend to more challenging settings, we also evaluate generalization to held-out locations, shapes, and sizes, non-retroreflective materials, and unseen object variations; full results are included in the supplement.

\newpage
\section{Applications}
\vspace{-0.5em}
\label{sec:capabilities}
We further demonstrate two key applications of \dsname{}: evaluating limitations of NLOS simulations and facilitating LiDAR design for NLOS perception.

\begin{figure}[t]
\centering
\includegraphics[width=\linewidth]{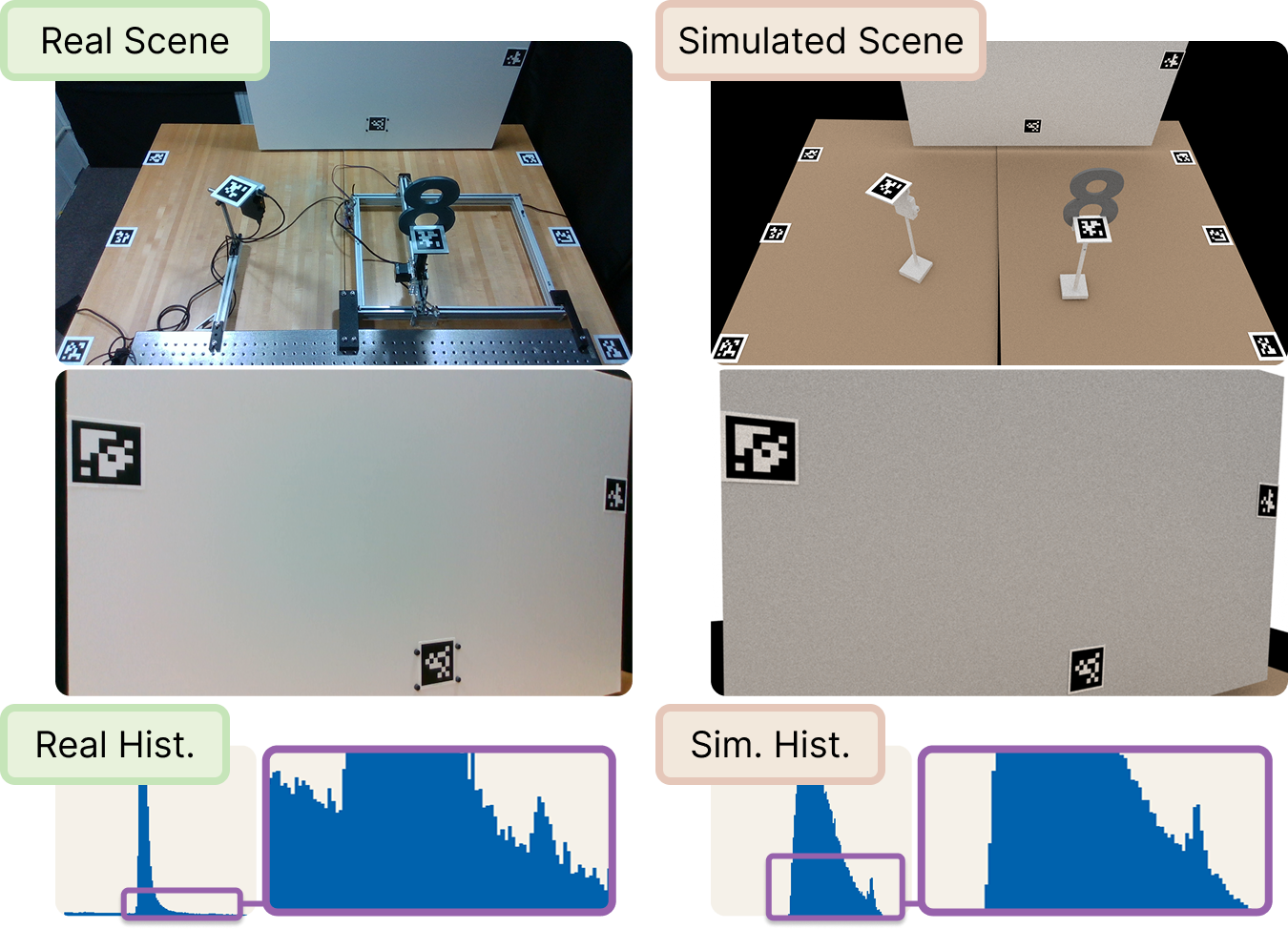}
\vspace{-6mm}
\caption{\textbf{Digital twins support evaluation of NLOS modeling fidelity.} For every dataset capture, we generate a scene digital twin in Mitsuba 3, enabling research on improving NLOS simulation and identifying key modeling discrepancies (e.g., pulse width, noise, jitter) that drive the sim-to-real gap for low-cost LiDARs.}
\vspace{-0.7em}
\label{fig:simulated_scene}
\end{figure}

\vspace{-0.1em}
\subsection{Evaluating Limitations of NLOS Simulations}
\label{sec:simulations}
\vspace{-0.2em}
We believe that our digital twins of captured scenes (\cref{subsec:poses}), rendered in Mitsuba 3 \cite{jakob2022mitsuba3}, provide a useful testbed for improving NLOS simulations for low-cost LiDARs. Each digital twin faithfully replicates our full capture setup, enabling LiDAR histograms to be rendered using existing simulation methods. In \cref{fig:simulated_scene}, we show an example histogram rendered with MiTransient \cite{royo2025mitransient}. While qualitatively similar, the simulated histogram fails to capture key effects such as true pulse width, noise characteristics, jitter, and intensity scaling. Full scene and NLOS simulation details are provided in the supplement.

\begin{figure}[!tb]
\centering
\includegraphics[width=\linewidth]{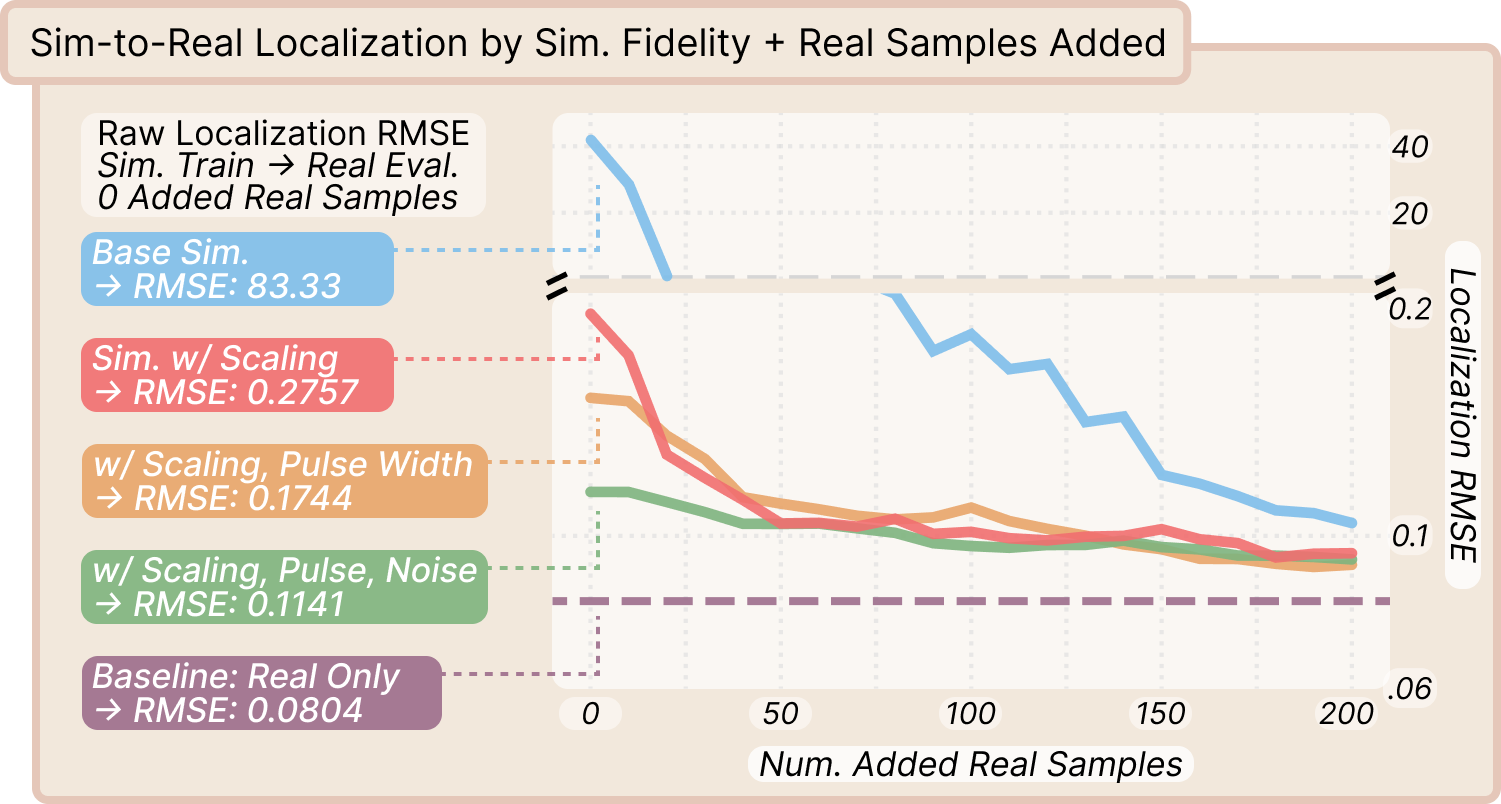}
\vspace{-1.5em}
\caption{\textbf{\dsname{} enables quantitative analysis of how modeled LiDAR effects impact sim-to-real transfer.} We train models on simulated center-pixel histograms from $3\times3$ light-on captures across object sizes, with increasing simulation fidelity (scaling, pulse width, noise), and evaluate on real measurements. Labels show sim-only RMSE; curves show smoothed ($w$=3) RMSE as real training samples are added. Higher simulation fidelity improves transfer with diminishing returns, while real data provides the largest gains at low fidelity.}
\label{fig:sim_to_real}
\vspace{-4mm}
\end{figure}

\bfheading{Evaluating NLOS Simulation Fidelity.} Using our digital twin scenes, we can identify which simulation factors are most critical for achieving robust sim-to-real transfer in low-cost LiDAR NLOS perception. Using MiTransient \cite{royo2025mitransient}, we render a simulated \textit{center-pixel histogram} for all $3\times3$ capture scenes and train a 1D CNN for NLOS object localization. To study key modeling factors affecting sim-to-real transfer, we learn three calibration functions for the simulated histograms: global scaling, pulse-width matching, and noise matching (details in the supplement). We then compare how localization RMSE improves when training on increasingly accurate simulations \textit{and} as real samples are added to the training set. Results are shown in \cref{fig:sim_to_real}. We find that \dsname{} provides a useful benchmark for evaluating simulation fidelity: it enables us to quantify how much localization error decreases for each simulation variant and to characterize the diminishing returns of adding real samples as simulation accuracy increases.

\subsection{Insights for Task-Specific Sensor Design}
\label{sec:sensor_des}
We also demonstrate how \dsname{} can inform task-specific sensor design for NLOS perception. We simulate increasing sensor temporal jitter by convolving histograms with Gaussian kernels of varying full width at half maximum (FWHM) during both training and evaluation. Results (\cref{tab:jitter_effect}) show that tasks exhibit different tolerances to timing blur, which can help identify minimum hardware requirements for a given NLOS application.

\begin{table}[ht]
\vspace{-0.5em}
\caption{\textbf{Effect of detector timing jitter} (Gaussian FWHM in ps) on downstream NLOS perception tasks, applied to $3\times3$ center-pixel histograms during training and evaluation.}
\vspace{-2mm}
\centering
\resizebox{\columnwidth}{!}{
\begin{tabular}{lccc}
\toprule
\multirow{2}{*}{\textbf{Time Jitter (ps)}} 
  & \textbf{Localization} & \textbf{Object Classification} & \textbf{Size Classification} \\
 & (RMSE$\downarrow$ / MAE$\downarrow$) & (Top1$\uparrow$ / Top5$\uparrow$ / M-F1$\uparrow$) & (P.$\uparrow$ / R.$\uparrow$ / Acc.$\uparrow$) \\
\midrule
0 (Baseline) 
  & 0.0804 / 0.0653 
  & 0.1554 / 0.4503 / 0.1416 
  & 0.8631 / 0.8616 / 0.8616 \\

$\sim$50\,ps 
  & 0.0802 / 0.0653 
  & 0.1525 / 0.4508 / 0.1374 
  & 0.8606 / 0.8599 / 0.8599 \\

$\sim$100\,ps 
  & 0.0802 / 0.0655 
  & 0.1684 / 0.4859 / 0.1560 
  & 0.8632 / 0.8616 / 0.8616 \\

$\sim$600\,ps 
  & 0.0819 / 0.0675 
  & 0.1260 / 0.4220 / 0.1064 
  & 0.8002 / 0.7944 / 0.7944 \\
\bottomrule
\end{tabular}}
\label{tab:jitter_effect}
\vspace{-2mm}
\end{table}

\vspace{-1em}
\section{Conclusion}
\label{sec:conclusion}

We introduce \dsname{}, the first large-scale dataset of time-resolved histograms from low-cost LiDARs for non-line-of-sight perception. Our aim is to make NLOS perception scalable, data-driven, and ultimately deployable on everyday LiDAR hardware (e.g., across phones and robotics). Just as large-scale benchmarks such as ImageNet accelerated progress in visual recognition, we believe that coupling modern learning methods with increasingly accessible dToF sensors can similarly advance \textit{data-driven NLOS perception}. \dsname{} represents a first step toward this goal of deployed NLOS perception on low-cost LiDAR systems.

\newpage 
\paragraph{Acknowledgements.} Nikhil Behari is supported by the NASA Space Technology Graduate Research Opportunity Fellowship. Suman Ghosh was supported in part by the NSF AccelNet-NeuroPAC Fellowship (OISE 2020624). We thank Aaron Young and Prof. Akshat Dave for their valuable contributions and insights supporting this work. We also thank Prof. Guillermo Gallego at Technische Universität Berlin for fostering collaboration through travel and institutional support.
{
    \small
    \bibliographystyle{ieeenat_fullname}
    \bibliography{main}
}



\end{document}